\documentclass[runningheads]{llncs}
\usepackage{graphicx}
\usepackage{times}
\usepackage{url}
\usepackage{latexsym}
\usepackage{xspace}
\usepackage{xAlgorithm}
\usepackage{multirow}
\usepackage{cite}
\usepackage{amsmath}
\usepackage{amssymb}
\usepackage{booktabs}
\usepackage{makecell}
\usepackage{ragged2e}
\usepackage{threeparttable}
\usepackage{color}
\usepackage[misc]{ifsym}

\let\vec\mathbf

\newcommand{\dbps}{\texttt{DBP15K}\xspace}
\newcommand{\dbpsz}{$\texttt{DBP15K}_\texttt{ZH-EN}$\xspace}
\newcommand{\dbpsj}{$\texttt{DBP15K}_\texttt{JA-EN}$\xspace}
\newcommand{\dbpsf}{$\texttt{DBP15K}_\texttt{FR-EN}$\xspace}

\newcommand{\bootea}{\textsf{BootEA}\xspace}
\newcommand{\gcnalign}{\textsf{GCN-Align}\xspace}

\newcommand{\hgcn}{\textsf{\small HGCN}\xspace}
\newcommand{\hgcnu}{\textsf{\small HGCN-U}\xspace}

\newcommand{\te}{\textsf{\small TransEdge}\xspace}
\newcommand{\mraea}{\textsf{\small MRAEA}\xspace}

\newcommand{\hman}{\textsf{\small HMAN}\xspace}
\newcommand{\imuse}{\textsf{\small IMUSE}\xspace}
\newcommand{\paris}{\textsf{\small PARIS}\xspace}
\newcommand{\ssp}{\textsf{\small SSP}\xspace}
\newcommand{\re}{\textsf{\small RE-GCN}\xspace}
\newcommand{\rrea}{\textsf{\small RREA}\xspace}
\newcommand{\dat}{\textsf{\small DAT}\xspace}

\newcommand{\our}{\textsf{\small UEA}\xspace}
\newcommand{\oura}{\textsf{\small UEA w/o Adj}\xspace}
\newcommand{\ouraa}{\textsf{\small UEA w/o Excl}\xspace}
\newcommand{\ourb}{\textsf{\small UEA w/o Prg}\xspace}

\newcommand{\oure}{\textsf{\small UEA-$\vec {M^l}$}\xspace}
\newcommand{\ourf}{\textsf{\small UEA-$\vec {M^n}$}\xspace}
\newcommand{\ourg}{\textsf{\small $\vec {M^n}$}\xspace}
\newcommand{\ourh}{\textsf{\small $\vec {M^l}$}\xspace}
\newcommand{\ouri}{\textsf{\small UEA w/o Unm}\xspace}

\newcommand{\transe}{\textsf{\small TransE}\xspace}
\newcommand{\gcn}{\textsf{\small GCN}\xspace}
\newcommand{\sota}{state-of-the-art\xspace}
\newtheorem{Example}{\textbf{Example}}
\newcommand{\myparagraph}[1]{\vspace{1ex}\noindent\textbf{#1.}\hspace{1em}}

\begin{document}
\title{Towards Entity Alignment in the Open World: \\An Unsupervised Approach}

\author{
	Weixin Zeng\inst{1}\and
	Xiang Zhao\inst{1} \Letter \and
	Jiuyang Tang\inst{1} \and
	Xinyi Li\inst{1} \and
	Minnan Luo\inst{2} \and
	Qinghua Zheng\inst{2}
}

\authorrunning{W. Zeng et al.}

%
\institute{Science and Technology on Information Systems Engineering Laboratory, National University of Defense Technology, Changsha, China \and
	Department of Computer Science, Xi'an Jiaotong University, Xi'an, China \\
	\email{\{zengweixin13,xiangzhao,jiuyang\_tang\}@nudt.edu.cn, lixinyimichael@gmail.com, \{minnluo,qhzheng\}@mail.xjtu.edu.cn}
}
\maketitle              
%
%
\begin{abstract}
Entity alignment (EA) aims to discover the equivalent entities in different knowledge graphs (KGs). It is a pivotal step for integrating KGs to increase knowledge coverage and quality. Recent years have witnessed a rapid increase of EA frameworks. However, state-of-the-art solutions tend to rely on labeled data for model training. Additionally, they work under the closed-domain setting and cannot deal with entities that are unmatchable. 

\quad\ To address these deficiencies, we offer an unsupervised framework that performs entity alignment in the open world. Specifically, we first mine useful features from the side information of KGs. Then, we devise an unmatchable entity prediction module to filter out unmatchable entities and produce preliminary alignment results. These preliminary results are regarded as the pseudo-labeled data and forwarded to the progressive learning framework to generate structural representations, which are integrated with the side information to provide a more comprehensive view for alignment. Finally, the progressive learning framework gradually improves the quality of structural embeddings and enhances the alignment performance by enriching the pseudo-labeled data with alignment results from the previous round. Our solution does not require labeled data and can effectively filter out unmatchable entities. Comprehensive experimental evaluations validate its superiority.

\end{abstract}
\section{Introduction}
\label{intro}
Knowledge graphs (KGs) have been applied to various fields such as natural language processing and information retrieval. 
To improve the quality of KGs, many efforts have been dedicated to the alignment of KGs, since different KGs usually contain complementary information. 
Particularly, entity alignment (EA), which aims to identify equivalent entities in different KGs, is a crucial step of KG alignment and has been intensively studied over the last few years~\cite{DBLP:conf/ccks/HaoZHL016,DBLP:conf/emnlp/ShiX19,DBLP:conf/emnlp/LiCHSLC19,DBLP:conf/aaai/SunW0CDZQ20,DBLP:conf/aaai/XuSFSY20,DBLP:conf/dasfaa/ChenGLZLZ20,DBLP:conf/acl/WuLFWZ20,DBLP:journals/pvldb/SunZHWCAL20}. 
We use Example~\ref{eg1} to illustrate this task. 

\begin{figure}[h]
	\centering
	\includegraphics[width=0.7\linewidth]{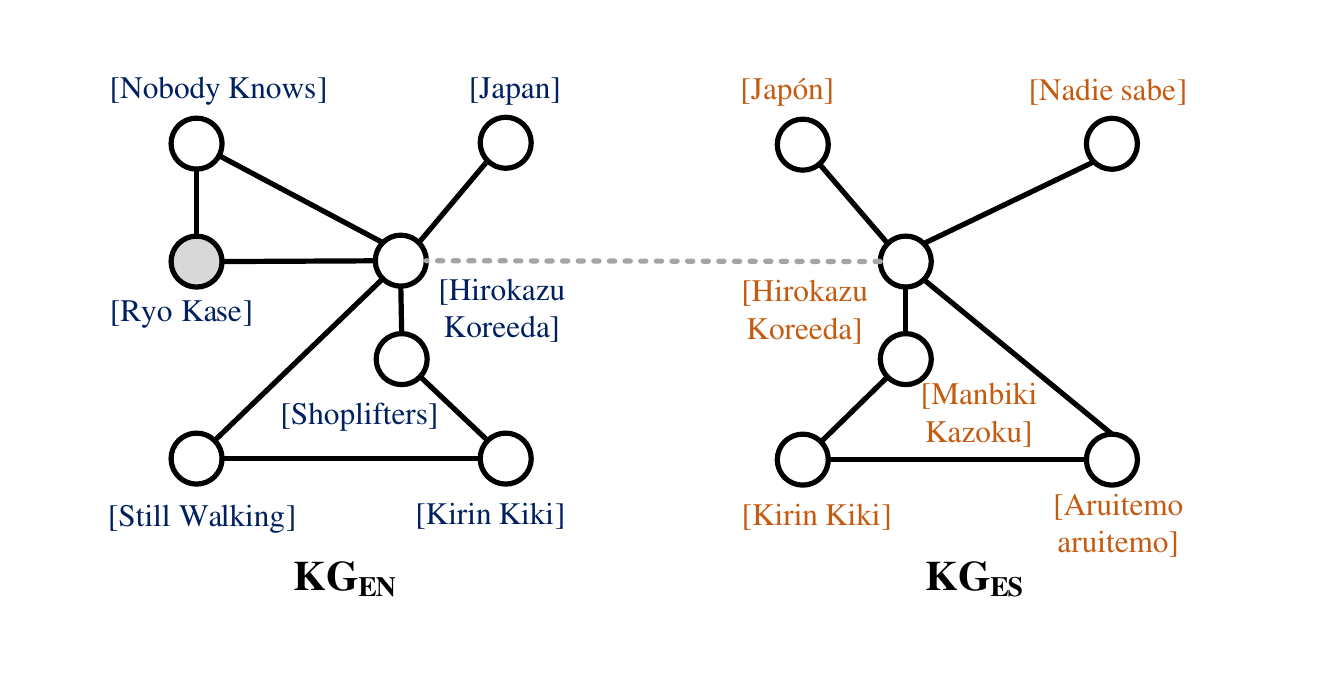}
	\caption{An example of EA.}
	\label{fig:ProblemDef}%
\end{figure}

\begin{Example}\label{eg1}
	In Figure~\ref{fig:ProblemDef} are a partial English KG and a partial Spanish KG concerning the director \texttt{Hirokazu Koreeda}, where the dashed lines indicate known alignments (i.e., seeds). The task of EA aims to identify equivalent entity pairs between two KGs, e.g., (\texttt{Shoplifters}, \texttt{Manbiki Kazoku}).
\end{Example} 

State-of-the-art EA solutions~\cite{DBLP:conf/ijcai/ChenTYZ17,DBLP:conf/semweb/SunHL17,DBLP:conf/emnlp/WuLFWZ19,cea} assume that equivalent entities usually possess similar neighboring information. Consequently, they utilize KG embedding models, e.g., \transe~\cite{DBLP:conf/nips/BordesUGWY13}, or graph neural network (GNN) models, e.g., \gcn~\cite{DBLP:journals/corr/KipfW16}, to generate structural embeddings of entities in individual KGs. 
Then, these separated embeddings are projected into a unified embedding space by using the seed entity pairs as connections, so that the entities from different KGs are directly comparable. 
Finally, to determine the alignment results, the majority of current works~\cite{DBLP:conf/semweb/SunHHCGQ19,DBLP:conf/emnlp/YangZSLLS19,DBLP:conf/emnlp/LiCHSLC19,DBLP:conf/acl/CaoLLLLC19} formalize the alignment process as a ranking problem; that is, for each entity in the source KG, they rank all the entities in the target KG according to some distance metric, and the closest entity is considered as the equivalent target entity. 

Nevertheless, we still observe several issues from current EA works:
\begin{itemize}
	\item \textbf{Reliance on labeled data}. Most of the approaches rely on pre-aligned seed entity pairs to connect two KGs and use the unified KG structural embeddings to align entities.
	These labeled data, however, might not exist in real-life settings. 
	For instance, in Example~\ref{eg1}, the equivalence between \texttt{Hirokazu Koreeda} in $KG_{EN}$ and \texttt{Hirokazu Koreeda} in $KG_{ES}$ might not be known in advance. 
	In this case, \sota methods that solely rely on the structural information would fall short, as there are no seeds to connect these individual KGs. 
	
	\item \textbf{Closed-domain setting}. All of current EA solutions work under the closed-domain setting~\cite{DBLP:conf/esws/HertlingP20}; that is, they assume every entity in the source KG has an equivalent entity in the target KG. 
	Nevertheless, in practical settings, there always exist unmatchable entities. 
	For instance, in Example~\ref{eg1}, for the source entity \texttt{Ryo Kase}, there is no equivalent entity in the target KG. 
	Therefore, an ideal EA system should be capable of predicting the unmatchable entities. 
\end{itemize}

In response to these issues, we put forward an unsupervised EA solution \our that is capable of addressing the unmatchable problem. 
Specifically, to mitigate the reliance on labeled data, we mine useful features from the KG side information and use them to produce preliminary pseudo-labeled data. 
These preliminary seeds are forwarded to our devised \textbf{progressive learning framework} to generate unified KG structural representations, which are integrated with the side information to provide a more comprehensive view for alignment. 
This framework also progressively augments the training data and improves the alignment results in a self-training fashion. 
Besides, to tackle the unmatchable issue, we design an \textbf{unmatchable entity prediction} module, which leverages thresholded bi-directional nearest neighbor search (TBNNS) to filter out the unmatchable entities and excludes them from the alignment results. 
We embed the unmatchable entity prediction module into the progressive learning framework to control the pace of progressive learning by dynamically adjusting the thresholds in TBNNS. 

\myparagraph{Contribution} The main contributions of the article can be summarized as follows:
\begin{itemize}
	\item We identify the deficiencies of existing EA methods, namely, requiring labeled data and working under the closed-domain setting, and propose an unsupervised EA framework \our that is able to deal with unmatchable entities. This is done by
	(1) exploiting the side information of KGs to generate preliminary pseudo-labeled data; 
	and (2) devising an unmatchable entity prediction module that leverages the thresholded bi-directional nearest neighbor search strategy to produce alignment results, which can effectively exclude unmatchable entities; 
	and (3) offering a progressive learning algorithm to improve the quality of KG embeddings and enhance the alignment performance. 
	
	\item We empirically evaluate our proposal against state-of-the-art methods, and the comparative results demonstrate the superiority of \our.
\end{itemize}

\myparagraph{Organization}
In Section~\ref{rel}, we formally define the task of EA and introduce related work. 
Section~\ref{me} elaborates the framework of \our.
In Section~\ref{exp}, we introduce experimental results and conduct detailed analysis. 
Section~\ref{con} concludes this article.

\section{Task Definition and Related Work}
\label{rel}
In this section, we formally define the task of EA, and then introduce the related work.

\myparagraph{Task definition}  
The inputs to EA are a source KG $G_1$ and a target KG $G_2$.
The task of EA is defined as finding the equivalent entities between the KGs, i.e., $\Psi = \{(u,v)|u\in E_1, v\in E_2, u \leftrightarrow v\}$, where $E_1$ and $E_2$ refer to the entity sets in $G_1$ and $G_2$, respectively, $u \leftrightarrow v$ represents the source entity $u$ and the target entity $v$ are \emph{equivalent}, i.e., $u$ and $v$ refer to the same real-world object. 

Most of current EA solutions assume that there exist a set of seed entity pairs $\Psi_s = \{(u_s,v_s)|u_s\in E_1, v_s\in E_2, u_s \leftrightarrow v_s\}$. 
Nevertheless, in this work, we focus on unsupervised EA and do not assume the availability of such labeled data. 

\myparagraph{Entity alignment} 
The majority of \sota methods are supervised or semi-supervised, which can be roughly divided into three categories, i.e., methods merely using the structural information, methods that utilize the iterative training strategy, and methods using information in addition to the structural information~\cite{9174835}. 

The approaches in the first category aim to mine useful structural signals for alignment, and devise structure learning models such as recurrent skipping networks~\cite{DBLP:conf/icml/GuoSH19} and multi-channel GNN~\cite{DBLP:conf/acl/CaoLLLLC19}, or exploit existing models such as \transe~\cite{DBLP:conf/ijcai/ChenTYZ17,DBLP:conf/emnlp/LiCHSLC19,DBLP:conf/ijcai/ZhuXLS17,DBLP:conf/ijcai/SunHZQ18,DBLP:conf/ijcai/ZhuZ0TG19} and graph attention networks~\cite{DBLP:conf/emnlp/LiCHSLC19}. 
The embedding spaces of different KGs are connected by seed entity pairs. 
In accordance to the distance in the unified embedding space, the alignment results can hence be predicted. 

Methods in the second category iteratively label likely EA pairs as the training set and gradually improve alignment results~\cite{DBLP:conf/ijcai/SunHZQ18,DBLP:conf/semweb/SunHHCGQ19,DBLP:conf/ijcai/ZhuZ0TG19,DBLP:conf/ijcai/ZhuXLS17,dat}. A more detailed discussion about these methods and the difference from our framework is provided in Section~\ref{prog}. 
Methods in the third category incorporate the side information to offer a complementing view to the KG structure, including the attributes~\cite{DBLP:conf/semweb/SunHL17,DBLP:conf/emnlp/WangLLZ18,DBLP:conf/aaai/TrisedyaQZ19,DBLP:conf/aaai/YangLZWX20,DBLP:conf/pakdd/ChenZT0L20,DBLP:conf/ijcai/Tang0C00L20}, entity descriptions~\cite{DBLP:conf/ijcai/ChenTCSZ18,DBLP:conf/emnlp/YangZSLLS19}, and entity names~\cite{ACL19,DBLP:conf/ijcai/WuLF0Y019,cea,dat,DBLP:conf/iclr/FeyL0MK20,zeng2021reinforcement}. 
These methods devise various models to encode the side information and consider them as features parallel to the structural information. 
In comparison, the side information in this work has an additional role, i.e., generating pseudo-labeled data for learning unified structural representations. 

\myparagraph{Unsupervised entity alignment} 
A few methods have investigated the alignment without labeled data. 
Qu et al.~\cite{DBLP:journals/corr/abs-1907-03179} propose an unsupervised approach towards knowledge graph alignment with the adversarial training framework. 
Nevertheless, the experimental results are extremely poor.  
He et al.~\cite{DBLP:conf/dasfaa/HeLQ0LZ0ZC19} utilize the shared attributes between heterogeneous KGs to generate aligned entity pairs, which are used to detect more equivalent attributes. They perform entity alignment and attribute alignment alternately, leading to more high-quality aligned entity pairs, which are used to train a relation embedding model. Finally, they combine the alignment results generated by attribute and relation triples using a bivariate regression model. 
The overall procedure of this work might seem similar to our proposed model. However, there are many notable differences; for instance, the KG embeddings in our work are updated progressively, which can lead to more accurate alignment results, and our model can deal with unmatchable entities. 
We empirically demonstrate the superiority of our model in Section~\ref{exp}.

We notice that there are some entity resolution (ER) approaches established in a setting similar to EA, represented by \paris~\cite{DBLP:journals/pvldb/SuchanekAS11}. 
They adopt collective alignment algorithms such as similarity propagation so as to model the relations among entities. We include them in the experimental study for the comprehensiveness of the article.

\section{Methodology}
\label{me}
In this section, we first introduce the outline of \our. Then, we elaborate its components. 

\begin{figure}[h]
	\centering
	\includegraphics[width=0.8\linewidth]{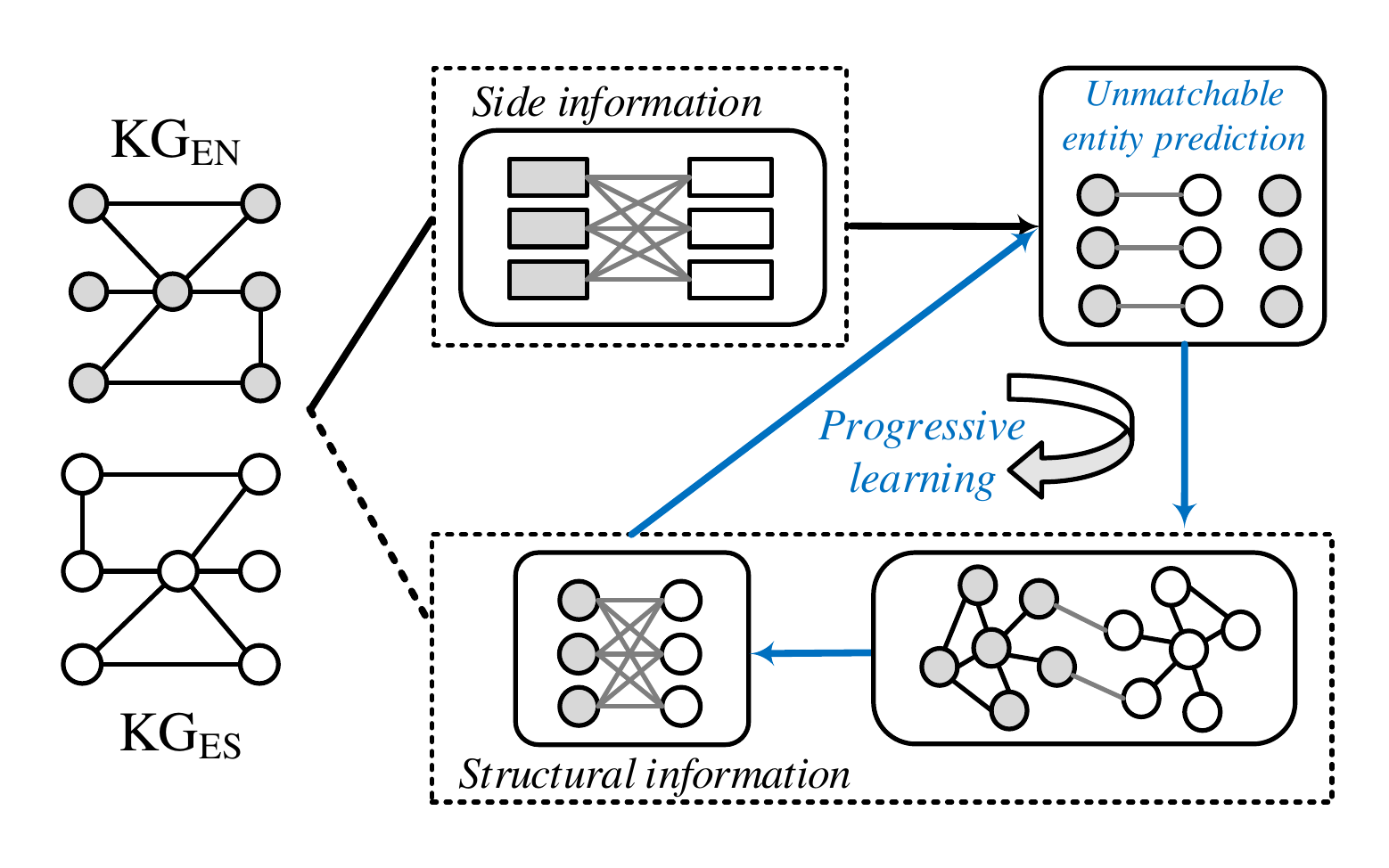}
	\caption{Outline of \our. Colored lines represent the progressive learning process.}
	\label{fig:overall}%
\end{figure}  

As shown in Figure~\ref{fig:overall}, given two KGs, \our first mines useful features from the \emph{side information}. These features are forwarded to the \emph{unmatchable entity prediction} module to generate initial alignment results, which are regarded as pseudo-labeled data.
Then, the \emph{progressive learning framework} uses these pseudo seeds to connect two KGs and learn unified entity structural embeddings. 
It further combines the alignment signals from the side information and \emph{structural information} to provide a more comprehensive view for alignment. 
Finally, it progressively improves the quality of structural embeddings and augments the alignment results by iteratively updating the pseudo-labeled data with results from the previous round, which also leads to increasingly better alignment.

\subsection{Side Information}
\label{side}
There is abundant side information in KGs, such as the attributes, descriptions and classes. 
In this work, we use a particular form of the attributes---the entity name, as it exists in the majority of KGs. 
To make the most of the entity name information, inspired by~\cite{cea}, we exploit it from the semantic level and string-level and generate the textual distance matrix between entities in two KGs.

More specifically, we use the averaged word embeddings to represent the semantic meanings of entity names. Given the semantic embeddings of a source and a target entity, we obtain the semantic distance score by subtracting their cosine similarity score from 1.
We denote the semantic distance matrix between the entities in two KGs as $\vec {M^n}$, where rows represent source entities, columns denote target entities and each element in the matrix denotes the distance score between a pair of source and target entities. 
As for the string-level feature, we adopt the Levenshtein distance~\cite{levenshtein1966binary} to measure the difference between two sequences. We denote the string distance matrix as $\vec {M^l}$. 

To obtain a more comprehensive view for alignment, we combine these two distance matrices and generate the textual distance matrix as $\vec {M^t} = \alpha\vec {M^n} + (1-\alpha)\vec {M^l}$, where $\alpha$ is a hyper-parameter that balances the weights. 
Then, we forward the textual distance matrix $\vec {M^t}$ into the unmatchable entity module to produce alignment results, which are considered as the pseudo-labeled data for training KG structural embeddings. The details are introduced in the next subsection. 

\myparagraph{Remark}
The goal of this step is to exploit available side information to generate useful features for alignment.  
Other types of side information, e.g., attributes and entity descriptions, can also be leveraged. 
Besides, more advanced textual encoders, such as misspelling oblivious word embeddings~\cite{DBLP:conf/naacl/EdizelPBFGS19} and convolutional embedding for edit distance~\cite{DBLP:conf/sigir/DaiYZW0C20}, can be utilized. 
We will investigate them in the future.

\subsection{Unmatchable Entity Prediction}
\label{thres}
State-of-the-art EA solutions generate for each source entity a corresponding target entity and fail to consider the potential unmatchable issue. 
Nevertheless, as mentioned in~\cite{9174835}, in real-life settings, KGs contain entities that other KGs do not contain. For instance, when aligning YAGO 4 and IMDB, only 1\% of entities in YAGO 4 are related to movies, while the other 99\% of entities in YAGO 4 necessarily have no match in IMDB. These unmatchable entities would increase the difficulty of EA. 
Therefore, in this work, we devise an unmatchable entity prediction module to predict the unmatchable entities and filter them out from the alignment results. 

More specifically, we put forward a novel strategy, i.e., thresholded bi-directional nearest neighbor search (TBNNS), to generate the alignment results, and the resulting unaligned entities are predicted to be unmatchable. 
As can be observed from Algorithm~\ref{alg:align}, given a source entity $u$ and a target entity $v$, if $u$ and $v$ are the nearest neighbor of each other, and the distance between them is below a given threshold $\theta$, we consider $(u,v)$ as an aligned entity pair. 
Note that $\vec M(u,v)$ represents the element in the $u$-th row and $v$-th column of the distance matrix $\vec M$.

\begin{algorithm}
	\Input{$G_1$ and $G_2$: the two KGs to be aligned; $E_1$ and $E_2$: the entity sets in $G_1$ and $G_2$; $\theta$: a given threshold; $\vec M$: a distance matrix.}
	\Output{$S$: Alignment results.}
	
	\ForEach{$u \in E_1$}{
		\State{$v \gets \mathop{\arg\min}\limits_{\hat v\in E_2} \vec M(u, \hat v)$}
		\If{$\mathop{\arg\min}\limits_{\hat u\in E_1} \vec M(v,\hat u) = u$ \AND $ \vec M(u,v) < \theta$}{
			{$S \gets S + \{(u,v)\}$
			}
		}
	}
	\Return{$S$.}
	\caption{TBNNS in the unmatchable entity prediction module } \label{alg:align}
\end{algorithm}

The TBNNS strategy exerts strong constraints on alignment, since it requires that the matched entities should both prefer each other the most, and the distance between their embeddings should be below a certain value. 
Therefore, it can effectively predict unmatchable entities and prevent them from being aligned. 
Notably, the threshold $\theta$ plays a significant role in this strategy. 
A larger threshold would lead to more matches, whereas it would also increase the risk of including erroneous matches or unmatchable entities. 
In contrast, a small threshold would only lead to a few aligned entity pairs, and almost all of them would be correct. 
This is further discussed and verified in Section~\ref{error}. 
Therefore, our progressive learning framework dynamically adjusts the threshold value to produce more accurate alignment results (to be discussed in the next subsection). 

\subsection{The Progressive Learning Framework}
\label{prog}
To exploit the rich structural patterns in KGs that could provide useful signals for alignment, we design a progressive learning framework to combine structural and textual features for alignment and improve the quality of both structural embeddings and alignment results in a self-training fashion. 

\myparagraph{Structural information}
As mentioned above, we forward the textual distance matrix $\vec {M^t}$ generated by using the side information to the unmatchable entity prediction module to produce the preliminary alignment results, which are considered as pseudo-labeled data for learning unified KG embeddings. 
Concretely, following~\cite{DBLP:conf/emnlp/WangLLZ18}, we adopt GCN~\footnote{More advanced structural learning models, such as recurrent skipping networks~\cite{DBLP:conf/icml/GuoSH19}, could also be used here. We will explore these alternative options in the future.} to capture the neighboring information of entities. We leave out the implementation details since this is not the focus of this paper, which can be found in~\cite{DBLP:conf/emnlp/WangLLZ18}.  

Given the learned structural embedding matrix $\vec Z$, we calculate the structural distance score between a source and a target entity by subtracting the cosine similarity score between their embeddings from 1. We denote the resultant structural distance matrix as $\vec {M^s}$. 
Then, we combine the textual and structural information to generate more accurate signals for alignment: $\vec {M} = \beta\vec {M^t} + (1-\beta)\vec {M^s}$, where $\beta$ is a hyper-parameter that balances the weights. 
The fused distance matrix $\vec {M}$ can be used to generate more accurate matches. 

\myparagraph{The progressive learning algorithm}
The amount of training data has an impact on the quality of the unified KG embeddings, which in turn affects the alignment performance~\cite{DBLP:conf/semweb/SunHL17,DBLP:conf/wsdm/MaoWXLW20}. 
As thus, we devise an algorithm (Algorithm~\ref{alg:all}) to progressively augment the pseudo training data, so as to improve the quality of KG embeddings and enhance the alignment performance. 
The algorithm starts with learning unified structural embeddings and generating the fused distance matrix $\vec {M}$ by using the preliminary pseudo-labeled data $S_0$ (line 1). 
Then, the fused distance matrix is used to produce the new alignment results $\Delta S$ using TBNNS (line 2). 
These newly generated entity pairs $\Delta S$ are added to the alignment results (which are considered as pseudo-labeled data for the next round), and the entities in the alignment results $S$ are removed from the entity sets (line 3-6).
In order to progressively improve the quality of KG embeddings and detect more alignment results, we perform the aforementioned process recursively until the number of newly generated entity pairs is below a given threshold $\gamma$ (line 7-13). 

\begin{algorithm}
	\Input{$G_1$ and $G_2$: the two KGs to be aligned; $E_1$ and $E_2$: the entity sets in $G_1$ and $G_2$; $\vec {M^t}$: textual distance matrix; $S_0$: preliminary labeled data; $\theta_0$: the initial threshold.}
	\Output{$S$: Alignment results.}
	\State{Use $S_0$ to learn structural embeddings, generate $\vec {M^s}$ and $\vec {M}$}
	\State{$\Delta S \gets$TBNNS($G_1$, $G_2$, $E_1$, $E_2$, $\theta_0$, $\vec {M}$)}
	\State{$S \gets S_0 + \Delta S$} 
	\State{$\theta \gets\theta_0+ \eta$}
	\State{$E_1 \gets \{e | e \in E_1, e \notin S\}$}
	\State{$E_2 \gets \{e | e \in E_2, e \notin S\}$}
	\While{the number of the newly generated alignment results is above $\gamma$}{
		\State{Use $S$ to learn structural embeddings, generate $\vec {M^s}$ and $\vec {M}$}
		\State{$\Delta S \gets$TBNNS($G_1$, $G_2$, $E_1$, $E_2$, $\theta$, $\vec {M}$)}
		\State{$S \gets S + \Delta S$}
		\State{$\theta \gets \theta$ + $\eta$}
		\State{$E_1 \gets \{e | e \in E_1, e \notin S\}$}
		\State{$E_2 \gets \{e | e \in E_2, e \notin S\}$}
	}
	\Return{$S$.}
	\caption{Progressive learning.} \label{alg:all}
\end{algorithm}

Notably, in the learning process, once a pair of entities is considered as a match, the entities will be removed from the entity sets (line 5-6 and line 12-13). 
This could gradually reduce the alignment search space and lower the difficulty for aligning the rest entities. 
Obviously, this strategy suffers from the error propagation issue, which, however, could be effectively mitigated by the progressive learning process that dynamically adjusts the threshold. 
We will verify the effectiveness of this setting in Section~\ref{abla}.

\myparagraph{Dynamic threshold adjustment}
It can be observed from Algorithm~\ref{alg:all} that, the matches generated by the unmatchable entity prediction module are not only part of the eventual alignment results, but also the pseudo training data for learning subsequent structural embeddings. 
Therefore, to enhance the overall alignment performance, the alignment results generated in each round should, ideally, have both large \emph{quantity} and high \emph{quality}. 
Unfortunately, these two goals cannot be achieved at the same time. 
This is because, as stated in Section~\ref{thres}, a larger threshold in TBNNS can generate more alignment results (large quantity), whereas some of them might be erroneous (low quality). These wrongly aligned entity pairs can cause the error propagation problem and result in more erroneous matches in the following rounds.  
In contrast, a smaller threshold leads to fewer alignment results (small quantity), while almost all of them are correct (high quality). 

To address this issue, we aim to balance between the quantity and the quality of the matches generated in each round. 
An intuitive idea is to set the threshold to a moderate value. 
However, this fails to take into account the characteristics of the progressive learning process. 
That is, in the beginning, the quality of the matches should be prioritized, as these alignment results will have a long-term impact on the subsequent rounds. 
In comparison, in the later stages where most of the entities have been aligned, the quantity is more important, as we need to include more possible matches that might not have a small distance score. 
In this connection, we set the initial threshold $\theta_0$ to a very small value so as to reduce potential errors. 
Then, in the following rounds, we gradually increase the threshold by $\eta$, so that more possible matches could be detected. 
We will empirically validate the superiority of this strategy over the fixed weight in Section~\ref{abla}.

\myparagraph{Remark}
As mentioned in the related work, there are some existing EA approaches that exploit the iterative learning (bootstrapping) strategy to improve EA performance. 
Particularly, \bootea calculates for each source entity the alignment likelihood to every target entity, and includes those with likelihood above a given threshold in a maximum likelihood matching process under the 1-to-1 mapping constraint, producing a solution containing confident EA pairs~\cite{DBLP:conf/ijcai/SunHZQ18}. This strategy is also adopted by~\cite{DBLP:conf/semweb/SunHHCGQ19,DBLP:conf/ijcai/ZhuZ0TG19}. 
Zhu et al. use a threshold to select the entity pairs with very close distances as the pseudo-labeled data~\cite{DBLP:conf/ijcai/ZhuXLS17}. 
\dat employs a bi-directional margin-based constraint to select the confident EA pairs as labels~\cite{dat}. 
Our progressive learning strategy differs from these existing solutions in three aspects: (1) we exclude the entities in the confident EA pairs from the test sets; and (2) we use the dynamic threshold adjustment strategy to control the pace of learning process; and (3) our strategy can deal with unmatchable entities. 
The superiority of our strategy is validated in Section~\ref{abla}.

\section{Experiment}
\label{exp}
This section reports the experiment results with in-depth analysis. The source code is available at \url{https://github.com/DexterZeng/UEA}. 

\subsection{Experiment Settings}
\label{setting}
\myparagraph{Datasets}
Following existing works, we adopt the \dbps dataset~\cite{DBLP:conf/semweb/SunHL17} for evaluation. 
This dataset consists of three multilingual KG pairs extracted from DBpedia. Each KG pair contains 15 thousand inter-language links as gold standards. The statistics can be found in Table~\ref{tab:data}. 
We note that \sota studies merely consider the labeled entities and divide them into training and testing sets. 
Nevertheless, as can be observed from Table~\ref{tab:data}, there exist unlabeled entities, e.g., 4,388 and 4,572 entities in the Chinese and English KG of \dbpsz, respectively. 
In this connection, we adapt the dataset by including the unmatchable entities. 
Specifically, for each KG pair, we keep 30\% of the labeled entity pairs as the training set (for training the supervised or semi-supervised methods). 
Then, to construct the test set, we include the rest of the entities in the first KG and the rest of the labeled entities in the second KG, so that the unlabeled entities in the first KG become unmatchable. 
The statistics of the test sets can be found in the \emph{Test set} column in Table~\ref{tab:data}.

\begin{table}[htbp]
	\small
	\centering
	\caption{The statistics of the evaluation benchmarks.}
	\begin{tabular}{l|lcccc|c}
		\toprule
		Dataset & KG pairs & \#Triples & \#Entities & \#Labeled Ents & \#Relations & \#Test set\\
		\midrule
		\multirow{2}[2]{*}{\dbpsz} & DBpedia(Chinese) & 70,414 & 19,388 & 15,000 & 1,701& 14,888\\
		& DBpedia(English) & 95,142 & 19,572 & 15,000 & 1,323 & 10,500\\
		\midrule
		\multirow{2}[2]{*}{\dbpsj} & DBpedia(Japanese) & 77,214 & 19,814 & 15,000 & 1,299& 15,314\\
		& DBpedia(English) & 93,484 & 19,780 & 15,000 & 1,153 & 10,500\\
		\midrule
		\multirow{2}[2]{*}{\dbpsf} & DBpedia(French) & 105,998 & 19,661 & 15,000 & 903 & 15,161\\
		& DBpedia(English) & 115,722 & 19,993 & 15,000 & 1,208 & 10,500\\
		\bottomrule
	\end{tabular}%
	\label{tab:data}%
\end{table}%

\myparagraph{Parameter settings}
For the {\itshape side information} module, we utilize the \textsf{fastText} embeddings~\cite{bojanowski2017enriching} as word embeddings. To deal with cross-lingual KG pairs, following~\cite{DBLP:conf/ijcai/WuLF0Y019}, we use Google translate to translate the entity names from one language to another, i.e., translating Chinese, Japanese and French to English. 
$\alpha$ is set to 0.5. 
For the {\itshape structural information learning}, we set $\beta$ to 0.5. 
Noteworthily, since there are no training set or validation set for parameter tuning, we set $\alpha$ and $\beta$ to the default value (0.5). 
We will further verify that the hyper-parameters do not have a large influence on the final results in Section~\ref{error}. 
For {\itshape progressive learning}, we set the initial threshold $\theta_0$ to 0.05, the incremental parameter $\eta$ to 0.1, the termination threshold $\gamma$ to 30. 
Note that if the threshold $\theta$ is over 0.45, we reset it to 0.45. 
These hyper-parameters are default values since there is no extra validation set for hyper-parameter tuning. 

\myparagraph{Evaluation metrics}
We use \emph{precision} (P), \emph{recall} (R) and \emph{F1 score} as evaluation metrics. 
The \emph{precision} is computed as the number of correct matches divided by the number of matches found by a method. 
The \emph{recall} is computed as the number of correct matches found by a method divided by the number of gold matches. 
The \emph{F1 score} is the harmonic mean between \emph{precision} and \emph{recall}. 

\myparagraph{Competitors}
We select the most performant \sota solutions for comparison. 
Within the group that solely utilizes structural information, we compare with \bootea~\cite{DBLP:conf/ijcai/SunHZQ18}, \te~\cite{DBLP:conf/semweb/SunHHCGQ19}, \mraea~\cite{DBLP:conf/wsdm/MaoWXLW20} and \ssp~\cite{DBLP:conf/ijcai/NieHSWCWZ20}.
Among the methods incorporating other sources of information, we compare with \gcnalign~\cite{DBLP:conf/emnlp/WangLLZ18}, \hman~\cite{DBLP:conf/emnlp/YangZSLLS19}, \hgcn~\cite{DBLP:conf/emnlp/WuLFWZ19}, \re~\cite{DBLP:conf/dasfaa/YangZWLHH20}, \dat~\cite{dat} and \rrea~\cite{DBLP:conf/cikm/MaoWXWL20}.
We also include the unsupervised approaches, i.e., \imuse~\cite{DBLP:conf/dasfaa/HeLQ0LZ0ZC19} and \paris~\cite{DBLP:journals/pvldb/SuchanekAS11}. 
To make a fair comparison, we only use entity name labels as the side information.

\begin{table}[htbp]
	\centering
	\caption{Alignment results.}
	\begin{tabular}{p{1.6cm}p{1.1cm}<{\centering}p{1.1cm}<{\centering}p{1.1cm}<{\centering}p{1.1cm}<{\centering}p{1.1cm}<{\centering}p{1.1cm}<{\centering}p{1.1cm}<{\centering}p{1.1cm}<{\centering}p{1.1cm}<{\centering}}
		\toprule
		\multirow{2}[4]{*}{} & \multicolumn{3}{c}{ZH-EN} & \multicolumn{3}{c}{JA-EN} & \multicolumn{3}{c}{FR-EN} \\
		\cmidrule{2-10}          & P     & R     & F1    & P     & R     & F1    & P     & R     & F1 \\
		\midrule
		\bootea & 0.444 & 0.629 & 0.520 & 0.426 & 0.622 & 0.506 & 0.452 & 0.653 & 0.534 \\
		\te & 0.518 & 0.735 & 0.608 & 0.493 & 0.719 & 0.585 & 0.492 & 0.710 & 0.581 \\
		\mraea & 0.534 & 0.757 & 0.626 & 0.520 & 0.758 & 0.617 & 0.540 & 0.780 & 0.638 \\
		\ssp   & 0.521 & 0.739 & 0.611 & 0.494 & 0.721 & 0.587 & 0.512 & 0.739 & 0.605 \\
		\midrule
		\gcnalign & 0.291 & 0.413 & 0.342 & 0.274 & 0.399 & 0.325 & 0.258 & 0.373 & 0.305 \\
		\hman  & 0.614 & 0.871 & 0.720 & 0.641 & \textbf{0.935} & 0.761 & 0.674 & \textbf{0.973} & 0.796 \\
		\hgcn  & 0.508 & 0.720 & 0.596 & 0.525 & 0.766 & 0.623 & 0.618 & 0.892 & 0.730 \\	
		\re & 0.518 & 0.735 & 0.608 & 0.548 & 0.799 & 0.650 & 0.646 & 0.933 & 0.764 \\
		\dat   & 0.556 & 0.788 & 0.652 & 0.573 & 0.835 & 0.679 & 0.639 & 0.922 & 0.755 \\
		\rrea  & 0.580 & 0.822 & 0.680 & 0.629 & 0.918 & 0.747 & 0.667 & 0.963 & 0.788 \\
		\midrule
		\imuse & 0.608 & 0.862 & 0.713 & 0.625 & 0.911 & 0.741 & 0.618 & 0.892 & 0.730 \\
		\paris & \textbf{0.976} & 0.777 & 0.865 & \textbf{0.981} & 0.785 & 0.872 & \textbf{0.972} & 0.793 & 0.873 \\
		\our   & 0.913 & \textbf{0.902} & \textbf{0.907} & 0.940 & 0.932 & \textbf{0.936} & 0.953 & 0.950 & \textbf{0.951} \\
		\bottomrule
	\end{tabular}%
	\label{tab:results}%
\end{table}%

\subsection{Results} \label{sec:results}
Table~\ref{tab:results} reports the alignment results, which shows that \sota supervised or semi-supervised methods have rather low precision values. 
This is because these approaches cannot predict the unmatchable source entities and generate a target entity for each source entity (including the unmatchable ones). 
Particularly, methods incorporating additional information attain relatively better performance than the methods in the first group, demonstrating the benefit of leveraging such additional information.

Regarding the unsupervised methods, although \imuse cannot deal with the unmatchable entities and achieves a low precision score, it outperforms most of the supervised or semi-supervised methods in terms of recall and F1 score. This indicates that, for the EA task, the KG side information is useful for mitigating the reliance on labeled data. 
In contrast to the abovementioned methods, \paris attains very high precision, since it only generates matches that it believes to be highly possible, which can effectively filter out the unmatchable entities. 
It also achieves the second best F1 score among all approaches, showcasing its effectiveness when the unmatchable entities are involved.
Our proposal, \our, achieves the best balance between precision and recall and attains the best F1 score, which outperforms the second-best method by a large margin, validating its effectiveness. 
Notably, although \our does not require labeled data, it achieves even better performance than the most performant supervised method \hman (except for the recall values on \dbpsj and \dbpsf). 

\subsection{Ablation Study}\label{abla}
In this subsection, we examine the usefulness of proposed modules by conducting the ablation study. 
More specifically, in Table~\ref{tab:abla}, we report the results of \ouri, which excludes the unmatchable entity prediction module, and \ourb, which excludes the progressive learning process. 
It shows that, removing the unmatchable entity prediction module (\ouri) brings down the performance on all metrics and datasets, validating its effectiveness of detecting the unmatchable entities and enhancing the overall alignment performance. 
Besides, without the progressive learning (\ourb), the precision increases, while the recall and F1 score values drop significantly. 
This shows that the progressive learning framework can discover more correct aligned entity pairs and is crucial to the alignment progress. 

To provide insights into the progressive learning framework, we report the results of \oura, which does not adjust the threshold, and 
\ouraa, which does not exclude the entities in the alignment results from the entity sets during the progressive learning. 
Table~\ref{tab:abla} shows that setting the threshold to a fixed value (\oura) leads to worse F1 results, verifying that the progressive learning process depends on the choice of the threshold and the quality of the alignment results. 
We will further discuss the setting of the threshold in the next subsection. 
Besides, the performance also decreases if we do not exclude the matched entities from the entity sets (\ouraa), validating that this strategy indeed can reduce the difficulty of aligning entities. 

Moreover, we replace our progressive learning framework with other \sota iterative learning strategies (i.e., \textsf{MWGM}~\cite{DBLP:conf/ijcai/SunHZQ18}, \textsf{TH}~\cite{DBLP:conf/ijcai/ZhuXLS17} and \textsf{DAT-I}~\cite{dat}) and report the results in Table~\ref{tab:abla}. 
It shows that using our progressive learning framework (\our) can attain the best F1 score, verifying its superiority. 

\begin{table}[htbp]
	\centering
	\caption{Ablation results.}
	\begin{tabular}{p{1.8cm}p{1.1cm}<{\centering}p{1.1cm}<{\centering}p{1.1cm}<{\centering}p{1.1cm}<{\centering}p{1.1cm}<{\centering}p{1.1cm}<{\centering}p{1.1cm}<{\centering}p{1.1cm}<{\centering}p{1.1cm}<{\centering}}
		\toprule
		\multirow{2}[4]{*}{} & \multicolumn{3}{c}{ZH-EN} & \multicolumn{3}{c}{JA-EN} & \multicolumn{3}{c}{FR-EN} \\
		\cmidrule{2-10}          & P     & R     & F1    & P     & R     & F1    & P     & R     & F1 \\
		\midrule
		\our   & 0.913 & 0.902 & \textbf{0.907} & 0.940 & 0.932 & \textbf{0.936} & 0.953 & 0.950 & \textbf{0.951} \\
		\textsf{\small w/o Unm} & 0.553 & 0.784 & 0.648 & 0.578 & 0.843 & 0.686 & 0.603 & 0.871 & 0.713 \\
		\textsf{\small w/o Prg} & 0.942 & 0.674 & 0.786 & 0.966 & 0.764 & 0.853 & 0.972 & 0.804 & 0.880 \\
		\textsf{\small w/o Adj} & 0.889 & 0.873 & 0.881 & 0.927 & 0.915 & 0.921 & 0.941 & 0.936 & 0.939 \\
		\textsf{\small w/o Excl} & \textbf{0.974} & 0.799 & 0.878 & 0.982 & 0.862 & 0.918 & 0.985 & 0.887 & 0.933 \\
		\midrule
		\textsf{MWGM}  & 0.930 & 0.789 & 0.853 & 0.954 & 0.858 & 0.903 & 0.959 & 0.909 & 0.934 \\
		\textsf{TH}    & 0.743 & \textbf{0.914} & 0.820 & 0.795 & \textbf{0.942} & 0.862 & 0.807 & \textbf{0.953} & 0.874 \\
		\textsf{DAT-I} & \textbf{0.974} & 0.805 & 0.881 & \textbf{0.985} & 0.866 & 0.922 & \textbf{0.988} & 0.875 & 0.928 \\
		\midrule
		\oure & 0.908 & 0.902 & 0.905 & 0.926 & 0.924 & 0.925 & 0.937 & 0.931 & 0.934 \\
		\ourh & 0.935 & 0.721 & 0.814 & 0.960 & 0.803 & 0.875 & 0.948 & 0.750 & 0.838 \\
		\ourf & 0.758 & 0.727 & 0.742 & 0.840 & 0.807 & 0.823 & 0.906 & 0.899 & 0.903 \\
		\ourg & 0.891 & 0.497 & 0.638 & 0.918 & 0.562 & 0.697 & 0.959 & 0.752 & 0.843 \\
		\bottomrule
	\end{tabular}%
	\label{tab:abla}%
\end{table}%

\subsection{Quantitative Analysis}\label{error}
In this subsection, we perform quantitative analysis of the modules in \our.

\begin{figure}[h]
	\centering
	\includegraphics[width=\linewidth]{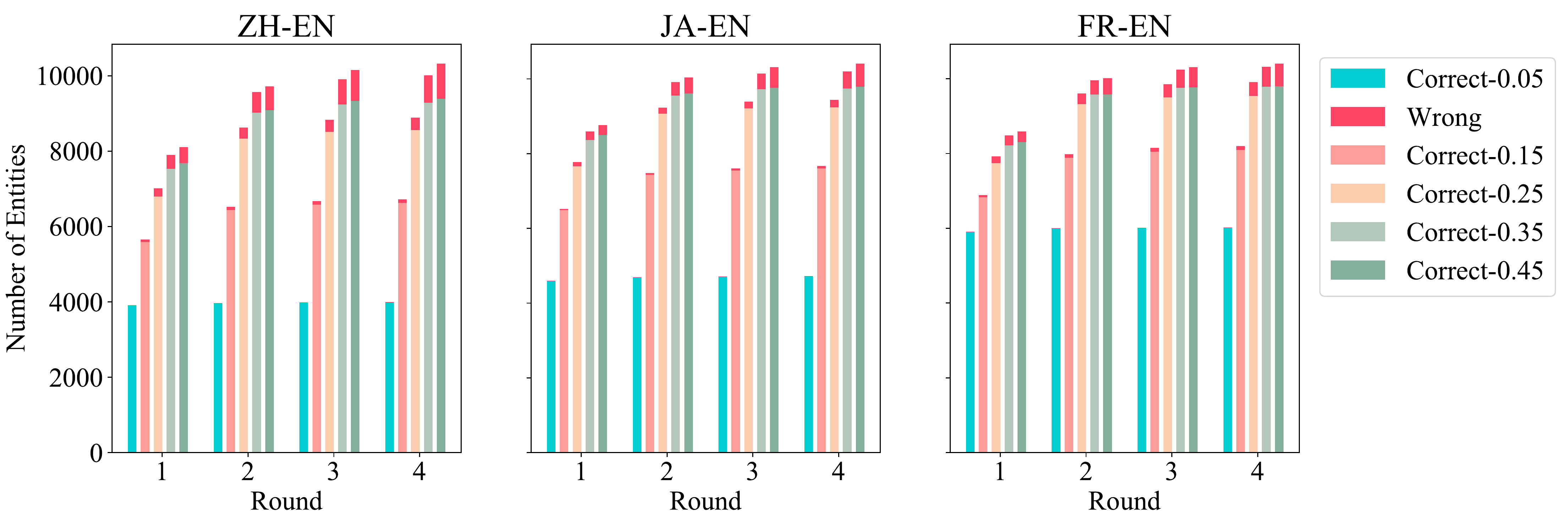}
	\caption{Alignment results given different threshold values. 
		Correct-$\theta$ refers to the number of correct matches generated by the progressive learning framework at each round given the threshold value $\theta$. Wrong refers to the number of erroneous matches generated in each round.} 
	\label{fig:ana}%
\end{figure}

\myparagraph{The threshold $\theta$ in TBNNS}
We discuss the setting of $\theta$ to reveal the trade-off between the risk and gain from generating the alignment results in the progressive learning.
Identifying a match leads to the integration of additional structural information, which benefits the subsequent learning.
However, for the same reason, the identification of a false positive, i.e., an incorrect match, potentially leads to mistakenly modifying the connections between KGs, with the risk of amplifying the error in successive rounds. 
As shown in Figure~\ref{fig:ana}, a smaller $\theta$ (e.g., 0.05) brings low risk and low gain; that is, it merely generates a small number of matches, among which almost all are correct. 
In contrast, a higher $\theta$ (e.g., 0.45) increases the risk, and brings relatively higher gain; that is, it results in much more aligned entity pairs, while a certain portion of them are erroneous. 
Additionally, using a higher threshold leads to increasingly more alignment results, while for a lower threshold, the progressive learning process barely increases the number of matches. 
This is in consistency with our theoretical analysis in Section~\ref{thres}.

\myparagraph{Unmatchable entity prediction}
Zhao et al.~\cite{9174835} propose an intuitive strategy (\textsf{U-TH}) to predict the unmatchable entities. They set an NIL threshold, and if the distance value between a source entity and its closest target entity is above this threshold, they consider the source entity to be unmatchable. 
We compare our unmatchable entity prediction strategy with it in terms of the percentage of unmatchable entities that are included in the final alignment results and the F1 score. 
On \dbpsz, replacing our unmatchable entity prediction strategy with \textsf{U-TH} attains the F1 score at 0.837, which is 8.4\% lower than that of \our.
Besides, in the alignment results generated by using \textsf{U-TH}, 18.9\% are unmatchable entities, while this figure for \our is merely 3.9\%. 
This demonstrates the superiority of our unmatchable entity prediction strategy. 

\begin{figure}[h]
	\centering
	\includegraphics[width=\linewidth]{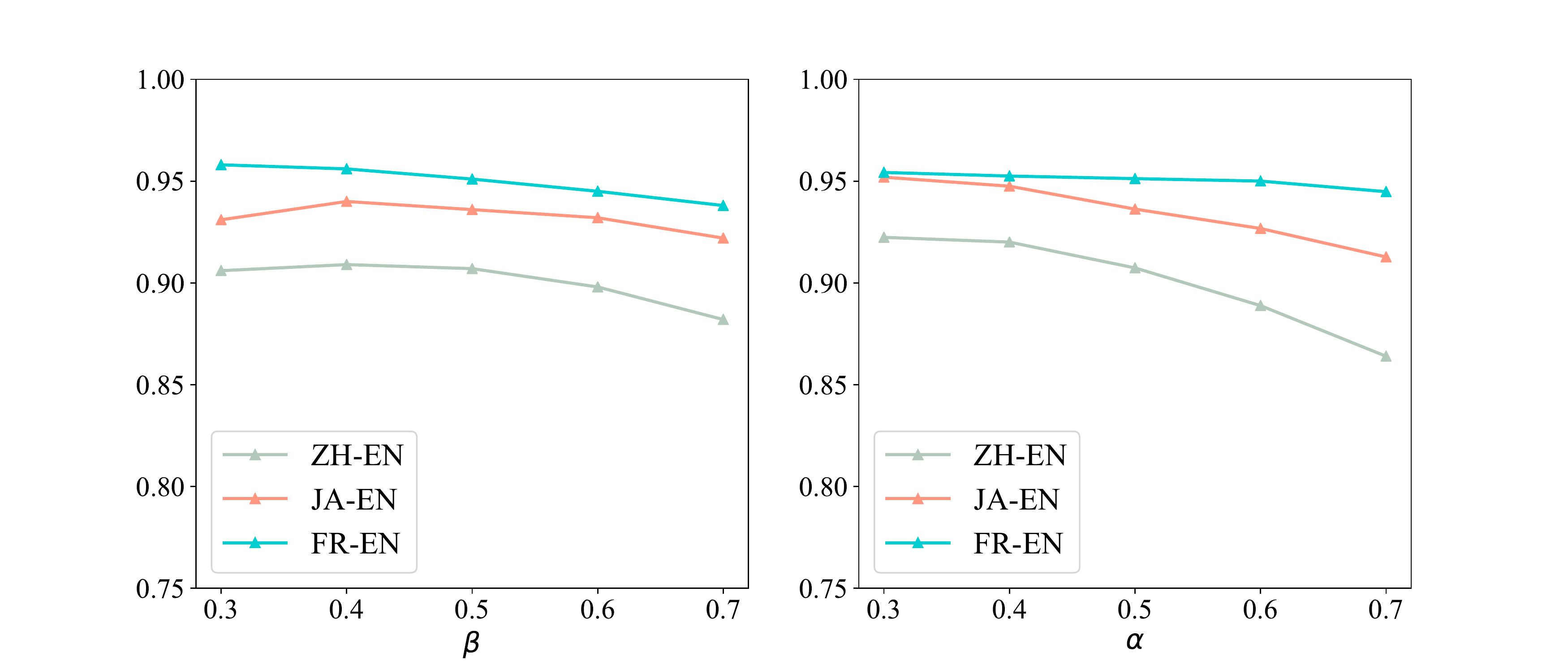}
	\caption{The F1 scores by setting $\alpha$ and $\beta$ to different values.}
	\label{fig:ana1}%
\end{figure}

\myparagraph{Influence of parameters}
As mentioned in Section~\ref{setting}, we set $\alpha$ and $\beta$ to 0.5 since there are no training/validation data. 
Here, we aim to prove that different values of the parameters do not have a large influence on the final results. 
More specifically, we keep $\alpha$ at 0.5, and choose $\beta$ from [0.3, 0.4, 0.5, 0.6, 0.7]; then we keep $\beta$ at 0.5, and choose $\alpha$ from [0.3, 0.4, 0.5, 0.6, 0.7]. 
It can be observed from Figure~\ref{fig:ana1} that, although smaller $\alpha$ and $\beta$ lead to better results, the performance does not change significantly. 

\myparagraph{Influence of input side information}
We adopt different side information as input to examine the performance of \our.
More specifically, we report the results of \oure, which merely uses the string-level feature of entity names as input, \ourf, which only uses the semantic embeddings of entity names as input. 
We also provide the results of \ourh and \ourg, which use the string-level and semantic information to directly generate alignment results (without progressive learning), respectively. 

As shown in Table~\ref{fig:ana}, the performance of solely using the input side information is not very promising (\ourh and \ourg). Nevertheless, by forwarding the side information into our model, the results of \oure and \ourf become much better. This unveils that \our can work with different types of side information and consistently improve the alignment results. 
Additionally, by comparing \oure with \ourf, it is evident that the input side information does affect the final results, and the quality of the side information is of significance to the overall alignment performance. 

\myparagraph{Pseudo-labeled data}
We further examine the usefulness of the preliminary alignment results generated by the side information, i.e., the pseudo-labeled data. 
Concretely, we replace the training data in \hgcn with these pseudo-labeled data, resulting in \hgcnu, and then compare its alignment results with the original performance. 
Regarding the F1 score, \hgcnu is 4\% lower than \hgcn on \dbpsz, 2.9\% lower on \dbpsj, 2.8\% lower on \dbpsf. 
The minor difference validates the effectiveness of the pseudo-labeled data generated by the side information. 
It also demonstrates that this strategy can be applied to other supervised or semi-supervised frameworks to reduce their reliance on labeled data.

\section{Conclusion}
\label{con}
In this article, we propose an unsupervised EA solution that is capable of dealing with unmatchable entities. 
We first exploit the side information of KGs to generate preliminary alignment results, which are considered as pseudo-labeled data and forwarded to the progressive learning framework to produce better KG embeddings and alignment results in a self-training fashion. 
We also devise an unmatchable entity prediction module to detect the unmatchable entities. 
The experimental results validate the usefulness of our proposed model and its superiority over \sota approaches. 
\\

\myparagraph{Acknowledgments} 
\footnotesize
This work was partially supported by 
Ministry of Science and Technology of China under grant No. 2020AAA0108800,
NSFC under grants Nos. 61872446 and 71971212, 
NSF of Hunan Province under grant No. 2019JJ20024, 
Postgraduate Scientific Research Innovation Project of Hunan Province under grant No. CX20190033. 

\bibliographystyle{unsrt}
\bibliography{dasfaa}

\end{document}